\documentclass[letterpaper, 10 pt, conference]{ieeeconf}  
\let\labelindent\relax

\usepackage{graphicx}
\usepackage{amsfonts}
\usepackage{amsmath,amssymb,bm}
\usepackage{upgreek}
\usepackage{multirow}
\usepackage{booktabs}
\usepackage{enumitem}
\usepackage{caption}
\usepackage{cite}

\IEEEoverridecommandlockouts             
\title{\LARGE \bf
StarIO: A Lightweight Inertial Odometry for Nonlinear Motion
}

\author{
Shanshan Zhang$^{1,2}$,  Siyue Wang$^{1}$,Qi Zhang$^{1}$,Liqin Wu$^{1}$, Tianshui Wen$^{1}$,  Ziheng Zhou$^{1}$, \\Xuemin Hong$^{1}$, Lingxiang Zheng$^{1}$, Yu Yang$^{2}$
\thanks{*This work is supported by Science and Technology Major Program of Fujian Province (No. 2022HZ026007) and partly supported by the Science and Technology Planning Project of Fujian province (2022I0001).}
\thanks{$^{1}$Shanshan Zhang, Qi Zhang, Siyue Wang, Tianshui Wen and Lingxiang Zheng are with the Department of Information and Communication Engineering, National and Local Joint Engineering Research Center of Navigation and Location-Based Services, Xiamen University, Xiamen 361005, China (e-mail: lxzheng@xmu.edu.cn).}
\thanks{Yu Yang and Shanshan Zhang are with the Department of Electronic Science, State Key Laboratory of Physical Chemistry of Solid Surfaces, Xiamen University, Xiamen 361005, China (e-mail: yuyang15@xmu.edu.cn).}
}

\begin{document}

\maketitle
\thispagestyle{empty}
\pagestyle{empty}
\begin{abstract}
Inertial odometry (IO) is an attractive approach for consumer-grade localization. However, existing data-driven IO methods often suffer from significant drift under complex nonlinear motion patterns (e.g., turns), as they struggle to capture the nonlinear relationships between Inertial Measurement Unit (IMU) signals and motion states. To address this issue, we propose a lightweight IO model, StarIO. Specifically, we first apply the Star Operation to project IMU signals into a high-dimensional implicit nonlinear feature space, enabling effective extraction of the complex nonlinear motion characteristics that typically cause drift. 
We then capture contextual dependencies across both the temporal and channel dimensions to enhance trajectory estimation over long sequences.
In addition, we introduce a multi-scale gated unit that fuses fine-grained local motion dynamics with contextual information to achieve a comprehensive representation of motion. 
Extensive experiments on six representative open-source datasets demonstrate that StarIO achieves a superior trade-off between model lightweightness and localization accuracy.
For example, on the RoNIN dataset, our approach reduces the ATE by 5.21\% compared to R-ResNet while using only 2.762M parameters. In order to comply with the double-blind review policy, the code will be made publicly available after the review process is completed.
\end{abstract}

\section{Introduction}
An inertial measurement unit (IMU) typically comprises a gyroscope and an accelerometer, providing high-rate measurements of acceleration and angular velocity. Inertial odometry (IO) estimates platform motion directly from IMU signals and is a key technology enabling the widespread adoption of consumer-grade positioning systems~\cite{RIO}. Because IO offers low-cost, stable, and robust motion estimation even under extreme or adverse conditions, it is indispensable for applications such as augmented reality, navigation for people with disabilities, and privacy-sensitive localization services~\cite{InformationAidedInertialNavigation,SurveyofIndoorInertial,CarIMU}.

Traditional IO derives motion estimates by integrating IMU signals based on Newtonian mechanics, but localization accuracy deteriorates rapidly due to error accumulation\cite{SINS}. Incorporating additional physical priors (e.g., motion constraints or dynamic models) can partially reduce drift, but such priors often restrict the range of applicable scenarios and motion patterns\cite{ZUPT,PDR,CHE,SC}. 
In recent years, data-driven IO has substantially improved localization accuracy and broadened the scope of feasible applications\cite{surveyILS}. However, these methods commonly exhibit significant drift in the presence of complex nonlinear motion patterns such as turns\cite{iMOT}. We attribute this limitation to the typical deployment of IO on edge devices, where constrained computational resources and storage capacity necessitate lightweight models for IO\cite{IMUNet}. Such models typically project IMU signals into a low-dimensional feature space with limited nonlinear representational capacity, making it difficult to model complex motion patterns. Therefore, a central challenge is to design a lightweight IO model that can efficiently project IMU signals—rich in motion information—into a high-dimensional nonlinear feature space so as to accurately capture complex nonlinear motion characteristics.

\begin{figure}[t]
\centering
\captionsetup{aboveskip=2pt,font=small}
\includegraphics[width=0.48\textwidth]{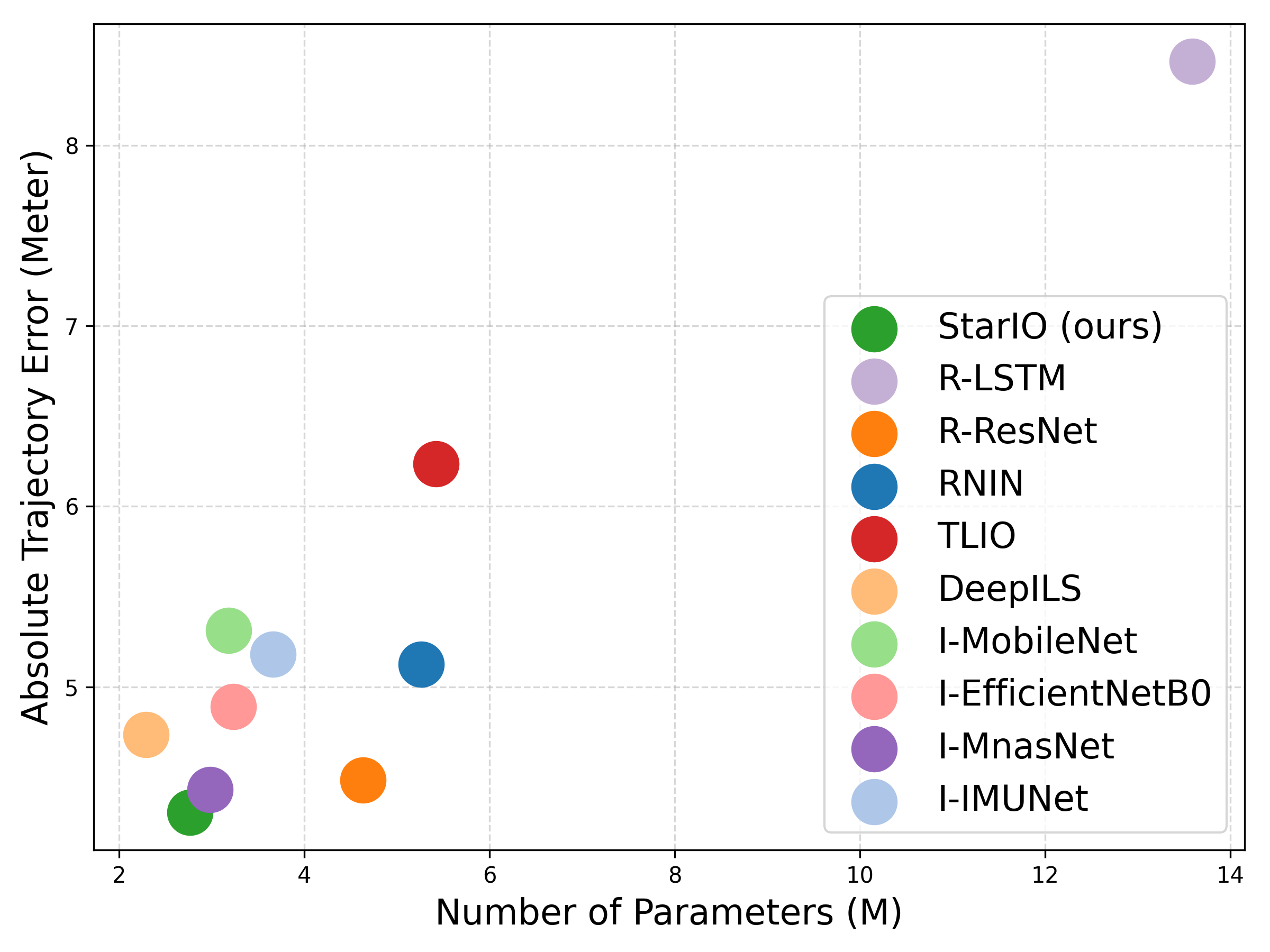}
\caption{Comparison of Absolute Trajectory Error (ATE) for different models on the RoNIN dataset. Points nearer the bottom-left indicate lower trajectory error and fewer parameters, reflecting a superior overall performance-efficiency trade-off.}
\label{ATE_vs_Params_TLIO}
\vspace{-24pt}
\end{figure}

To address these challenges, this paper proposes StarIO, a lightweight IO model that incorporates the Star Operation \cite{StarNet}. Specifically, the Star Operation applies element-wise self-products on extracted features, thereby expanding the feature space and introducing nonlinear representations without increasing network depth.
This projection facilitates subsequent extraction of rich contextual features and fusion of multi-scale motion characteristics. The main contributions of this work are as follows:
\begin{itemize}[noitemsep, nolistsep, leftmargin=*]
    \item We propose using the Star Operation to efficiently project IMU signals into an implicit high-dimensional nonlinear feature space, substantially enhancing the capacity to model complex nonlinear motion characteristics.
    \item We introduce a lightweight dual-wing attention mechanism that models contextual motion information along both the channel and temporal dimensions within the implicit high-dimensional feature space.
    \item We design a multi-scale gated unit that effectively aggregates fine-grained local motion dynamics with contextual information to form a comprehensive motion representation.
    \item Extensive experiments on multiple open-source datasets demonstrate that StarIO achieves an excellent trade-off between localization accuracy and model size, as illustrated in Fig.~\ref{ATE_vs_Params_TLIO}.
\end{itemize}

\begin{figure*}[!ht]
\centering
\captionsetup{aboveskip=2pt,font=small}
\includegraphics[width=1.0\textwidth]{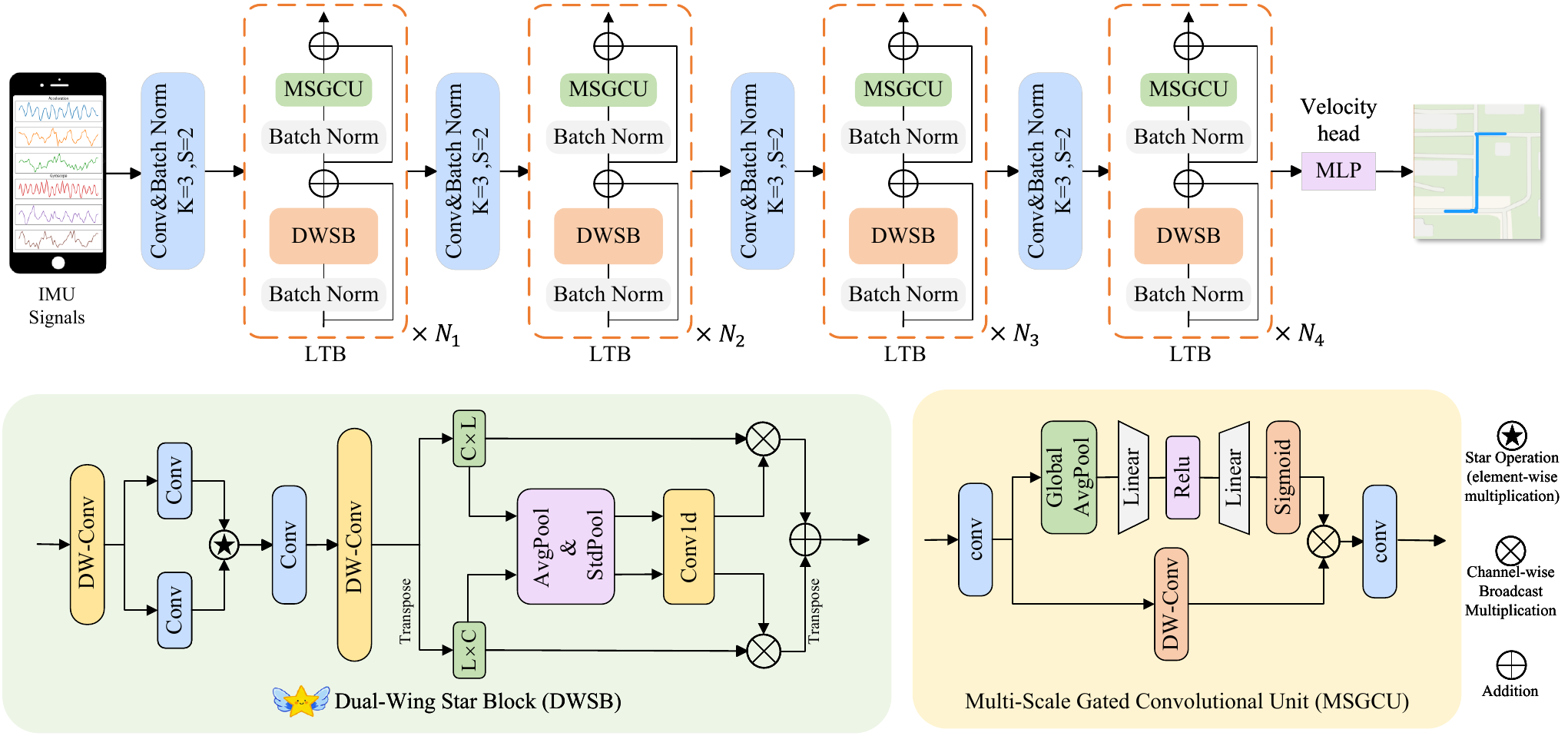}
\caption{The overall architecture of the proposed StarIO, which primarily consists of the Lightweight Transform Block (LTB) featuring the Dual-Wing Star Block (DWSB) and Multi-Scale Gated Convolution Unit (MSGCU).}
\label{framework}
\vspace{-15pt}
\end{figure*}

\section{Related Work}
\subsection{Newtonian mechanics-based IO}
Traditional IO relies on Newtonian kinematic principles to directly compute motion information\cite{surveyILS}. These methods do not require large-scale IMU datasets or significant computational resources, but they are often limited in accuracy and robustness. For example, strapdown inertial navigation systems (SINS) use high-precision and costly IMUs to compute displacement via double integration~\cite{SINS,SINS-base-dual-quaternions,IMU-WIFI}; however, persistent error accumulation renders long-range localization unreliable. To mitigate error accumulation, researchers have introduced physical priors to correct such errors. Pedestrian dead reckoning (PDR) assumes a regular walking pattern~\cite{PDR}. The zero-velocity update (ZUPT) technique is applied by mounting the IMU on the foot and enforcing zero-velocity constraints during stance phases to reduce system errors~\cite{ZUPT}. Coriolis-based heading estimation (CHE) builds on ZUPT by incorporating a magnetometer and exploiting the Coriolis effect for heading estimation to correct residual errors\cite{CHE}. However, the introduction of physical priors often relies on strong assumptions about motion patterns, which limits the generalization and robustness of IO methods in complex motion scenarios~\cite{RoNIN}.

\subsection{Data-driven IO}
The recent introduction of machine learning approaches has further improved IO localization accuracy and expanded its applicability\cite{SurveyofIndoorInertial,AirIO,Tartan-IMU,IDOL,PDRTransform}. RIDI~\cite{RIDI} and PDRNet~\cite{PDRNet} first identify IMU placement and then apply dedicated network architectures for position inference. IONet~\cite{IONet} and RoNIN~\cite{RoNIN} are agnostic to IMU placement and instead use unified models to infer motion information directly from IMU signals, demonstrating a degree of robustness.

Building on the ResNet architecture proposed in RoNIN, TLIO\cite{TLIO} and LIDR\cite{LIDR} introduce the State-Cloning Extended Kalman Filter (SCEKF) and the Left-Invariant Extended Kalman Filter (LIEKF), respectively, to further reduce localization error. Similarly, RNIN\cite{RNIN-VIO}, SCHNN\cite{SCHNN}, SSHNN\cite{SSHNN}, and CTIN\cite{CTIN} combine convolutional neural networks with LSTMs or attention mechanisms to capture contextual motion information. RIO\cite{RIO} and EqNIO\cite{EqNIO} design modular network components to address orientation bias in IMU signals.

To facilitate mobile deployment, IMUNet\cite{IMUNet} adopts depthwise separable convolutions to create a lightweight IO, and DeepILS\cite{DeepILS} further incorporates CBAM\cite{PAMCAM} to enhance contextual motion awareness. Because IO is often deployed on mobile devices, data-driven IO methods tend to favor shallow architectures with small parameter counts. However, shallow networks typically struggle to map IMU signals into a high-dimensional nonlinear feature space capable of modeling complex nonlinear characteristics. Real-world IMU data are often highly intricate, which further challenges the design of effective models. Therefore, designing lightweight IO architectures that efficiently project IMU signals into high-dimensional nonlinear feature spaces and accurately model their complex nonlinear characteristics remains an important research challenge.

\section{Proposed Method}
In this section, we present the overall pipeline of StarIO and then describe its core component—the Lightweight Transform Block (LTB), which comprises two submodules: the Dual-Wing Star Block (DWSB) and the Multi-Scale Gated Convolution Unit (MSGCU).

\subsection{Overall Architecture}
As illustrated in Fig.~\ref{framework}, StarIO employs a Transformer-based encoder. Let \(\mathbf{X}\in\mathbb{R}^{C\times L}\) denote IMU signals collected over a temporal window, where \(C=6\) corresponds to the three-axis angular velocities from the gyroscope and the three-axis accelerations from the accelerometer, and \(L\) is the number of samples in the window.

We first apply a one-dimensional convolutional layer with kernel size \(3\) along the temporal axis to adjust the channel dimension of \(\mathbf{X}\). Within the main backbone, the network comprises four stages indexed by \(i\in\{1,2,3,4\}\). Stage \(i\) stacks \(N_i\) LTBs to progressively capture complex, global motion dynamics. Downsampling between stages is performed using one-dimensional convolutions with stride of \(3\).

Within each LTB, the DWSB applies the Star Operation~\cite{StarNet} to project IMU signals into a high-dimensional, nonlinear feature space, enhancing representations of complex motion patterns and capturing global dynamics across both the channel and temporal dimensions. The output of the DWSB is then fed into the MSGCU, which fuses local fine-grained motion features with cross-channel dependencies, producing richer and more diverse motion representations.

Finally, StarIO maps features to the average velocity within the window via a multi-layer perceptron (MLP) and obtains displacement by temporal integration. We train the model with the mean squared error (MSE) loss.

\subsection{Lightweight Transform Block}
The Lightweight Transform Block (LTB) is the core module of StarIO and consists of two submodules: the DWSB and the MSGCU. To avoid the computationally expensive standard attention mechanism, we employ an efficient nonlinear feature-space mapping to achieve lightweight IO.

Let \(\mathbf{X}_i\) denote the input features to the \(i\)-th LTB. The processing within an LTB is expressed as
\begin{equation}
  \mathbf{X}_{i}^{\prime} = \mathbf{X}_i + \text{DWSB}\bigl(\mathbf{X}_i\bigr),
\end{equation}
\begin{equation}
  \mathbf{X}_{i+1} = \mathbf{X}_{i}^{\prime} + \text{MSGCU}\bigl(\mathbf{X}_{i}^{\prime}\bigr).
\end{equation}
Here, \(\mathbf{X}_{i}^{\prime}\) and \(\mathbf{X}_{i+1}\) are the outputs of the DWSB and MSGCU modules, respectively.

\textbf{Dual-Wing Star Block (DWSB).}
In IO methods, deep neural networks typically rely on convolutional or fully connected layers for feature transformation, coupled with nonlinear activation functions to enhance model expressiveness. However, such approaches exhibit limitations in modeling high-dimensional nonlinear representations. 
To address these limitations, we propose the Dual-Wing Star Block (DWSB), which combines the high-dimensional nonlinear mapping capacity of the Star Operation with a lightweight dual-wing attention mechanism for the efficient modeling of complex motion features and global motion dynamics in IMU signal data.

\textit{1) High-Dimensional Nonlinear Mapping.} 
IMU signals are inherently nonlinear and dynamic, requiring strong nonlinear representational capacity from the model. 
To effectively extract and model complex motion features, we first project the IMU signals into a high-dimensional implicit nonlinear feature space using the Star Operation.
To simplify the derivation, we consider the IMU signal at a single time step as $\mathbf{X_{t}} \in \mathbb{R}^{C \times 1}$. Its linear transformation can be formulated as:
\begin{equation}
\mathbf{W'}_{1} \mathbf{X_{t}'} = \mathbf{W}_{1} \mathbf{X_{t}} + \mathbf{B}_{1}
\end{equation}
where $\mathbf{W}_{1} \in \mathbb{R}^{M \times C}$ is the weight matrix, $\mathbf{B}_{1} \in \mathbb{R}^{M \times 1}$ is the bias, $\mathbf{W'}_{1} = [\mathbf{W}_{1}, \mathbf{B}_{1}] \in \mathbb{R}^{M \times (C+1)} $, and $\mathbf{X_{t}'} = [\mathbf{X_{t}}; 1] \in \mathbb{R}^{(C+1) \times 1}$. Although this transformation increases the dimensionality when \( M > C \), the nonlinear capacity still largely depends on the subsequent activation functions.

To enhance nonlinear representational capacity, we introduce the Star Operation, which facilitates feature interactions via element-wise multiplication. This enables an efficient mapping to a high-dimensional implicit nonlinear space, which is equivalent to implicit feature dimensionality expansion.

Specifically, we first apply a depthwise convolution to extract local features for each channel. Let the output of the depthwise convolution at a specific time step $t$ be $\mathbf{X_{t}} \in \mathbb{R}^{C \times 1}$, where $C$ is the number of channels. Then, two parallel pointwise convolution branches are employed to capture inter-channel contextual information. Their outputs are linearly transformed as follows:
\begin{equation}
\mathbf{W'}_{2} \mathbf{X_{t}'} = \mathbf{W}_{2} \mathbf{X_{t}} + \mathbf{B}_{2}
\end{equation}
\begin{equation}
\mathbf{W'}_{3} \mathbf{X_{t}'} = \mathbf{W}_{3} \mathbf{X_{t}} + \mathbf{B}_{3}
\end{equation}
where $\mathbf{X_{t}'} = [\mathbf{X_{t}}; 1] \in \mathbb{R}^{(C+1) \times 1}$ is the augmented feature vector. Here, $\mathbf{W}_{2}, \mathbf{W}_{3} \in \mathbb{R}^{M \times C}$ and $\mathbf{B}_{2}, \mathbf{B}_{3} \in \mathbb{R}^{M \times 1}$ denote the weight matrices and bias terms, respectively. We define the augmented weight matrices as $\mathbf{W'}_{2} = [\mathbf{W}_{2}, \mathbf{B}_{2}]$ and $\mathbf{W'}_{3} = [\mathbf{W}_{3}, \mathbf{B}_{3}]$, where $\mathbf{W'}_{2}, \mathbf{W'}_{3} \in \mathbb{R}^{M \times (C+1)}$.

Instead of simple addition, we apply the Star Operation (element-wise multiplication)\cite{StarNet} to project the features into a high-dimensional implicit nonlinear space:
\begin{equation}
\begin{aligned}
&(\mathbf{W'}_{2} \mathbf{X_{t}'}) \ast (\mathbf{W'}_{3} \mathbf{X_{t}'})\\
&= \left(\sum_{i=1}^{M} w_{2,i}' x_{i}'\right) * \left(\sum_{j=1}^{M} w_{3,j}' x_{j}'\right) \\
&= \sum_{i=1}^{M} \sum_{j=1}^{M} w_{2,i}' w_{3,j}' x_{i}' x_{j}' \\
&= \underbrace{\beta_{(1,1)} x_{1}' x_{1}' + \cdots  \beta_{(4,5)} x_{4}' x_{5}' + \cdots  \beta_{(M,M)} x_{M}' x_{M}'}_{M(M+1)/2 \text{ terms}}
\end{aligned}
\end{equation}

\begin{table*}[!t]
\centering
\captionsetup{aboveskip=2pt,font=small}
\scriptsize
\renewcommand{\arraystretch}{0.9}
\caption{Trajectory Prediction Accuracy of StarIO and Baseline Models. The best results are highlighted in bold.}
\label{results}
\resizebox{\textwidth}{!}{
\begin{tabular}{llccccccccccc}
\toprule
\multirow{3}{*}{\textbf{Dataset}} & \multirow{3}{*}{\textbf{Metric}} 
& \textbf{R-ResNet} & \textbf{R-LSTM} & \textbf{RNIN} & \textbf{I-IMUNet} 
& \textbf{I-MobileNet} & \textbf{I-MnasNet} & \textbf{I-EfficientNetB0} & \textbf{DeepILS} 
& \textbf{TLIO} & \textbf{StarIO} 
& \multirow{3}{*}{\parbox{2.5cm}{\centering \textbf{StarIO Improvement over R-ResNet}}} \\ 

\cmidrule(lr){3-12}
& & \multicolumn{10}{c}{\textbf{Performance (meters)}} & \\ 
\midrule

\multirow{2}{*}{RoNIN}
& ATE & 4.165 & 8.103 & 4.781 & 4.755 & 4.883 & 4.077 & 4.629 & 4.374 & 5.805 & \textbf{3.948} & 5.210\% \\
& RTE & 3.724 & 4.836 & 3.705 & 3.736 & 3.713 & 3.572 & 3.681 & 3.665 & 4.276 & \textbf{3.554} & 4.565\% \\

\multirow{2}{*}{RIDI}
& ATE & 2.080 & 3.839 & 2.103 & 2.486 & 2.500 & 2.290 & 2.379 & 2.047 & 2.563 & \textbf{2.033} & 2.260\% \\
& RTE & 2.383 & 3.589 & 2.445 & 2.644 & 2.584 & 2.573 & 2.406 & 2.395 & 2.929 & \textbf{2.274} & 4.574\% \\

\multirow{2}{*}{RNIN}
& ATE & 1.844 & 3.182 & 1.605 & 2.174 & 3.119 & 3.142 & 2.648 & 2.478 & 2.428 & \textbf{1.455} & 21.095\% \\
& RTE & 1.836 & 3.726 & 1.593 & 2.114 & 1.568 & 1.842 & 1.613 & 1.499 & 1.979 & \textbf{1.370} & 25.381\% \\

\multirow{2}{*}{IMUNet}
& ATE & 6.428 & 8.108 & 6.445 & 9.130 & 6.846 & 8.593 & 6.479 & 6.918 & 10.240 & \textbf{4.964} & 22.775\% \\
& RTE & 4.802 & 6.381 & 4.463 & 6.404 & 4.809 & 5.609 & 4.458 & 5.638 & 5.735 & \textbf{3.815} & 20.554\% \\

\multirow{2}{*}{TLIO}
& ATE & 1.333 & 2.145 & 1.300 & 1.354 & 1.369 & 1.373 & 1.724 & 1.258 & 1.832 & \textbf{1.108} & 16.879\% \\
& RTE & 3.705 & 3.685 & 3.627 & 4.060 & 0.842 &\textbf{0.806} & 0.959 & 0.818 & 5.162 & 3.145 & 15.115\% \\

\multirow{2}{*}{OxIOD}
& ATE & 4.336 & 7.282 & 2.780 & 2.325 & 2.381 & 2.163 & 1.982 & 1.696 & 3.855 & \textbf{1.484} & 65.775\% \\
& RTE & 1.595 & 2.301 & 1.082 & 1.135 & 1.142 & 1.103 & 0.979 & 1.100 & 1.270 & \textbf{0.856} & 46.332\% \\
\bottomrule
\noalign{\vskip 2pt}  
\multicolumn{2}{l}{\textbf{Parameters (M)}} 
& 4.635 & 13.589 & 5.259 & 3.659 & 3.179 & 2.979 & 3.232 & \textbf{2.291} & 5.424 & 2.762 & 40.410\% \\

\multicolumn{2}{l}{\textbf{FLOPs (M)}} 
& 38.252 & 57.473 & 72.032 & 18.838 & 34.559 & 25.256 & 28.831 & \textbf{15.287} & 39.436 & 25.122 & 34.325\% \\

\multicolumn{2}{l}{\textbf{Peak Memory (MB)}} 
& 22.835 & 70.755 & 25.490 & 38.274 & 67.257 &\textbf{13.073} & 25.631 & 17.746 & 60.026 & 20.458 & 10.409\% \\
\bottomrule
\end{tabular}
}
\vspace{-10pt}
\end{table*}

\begin{figure*}[!t]
\centering
\captionsetup{aboveskip=2pt,font=small}
\includegraphics[width=1.0\textwidth]{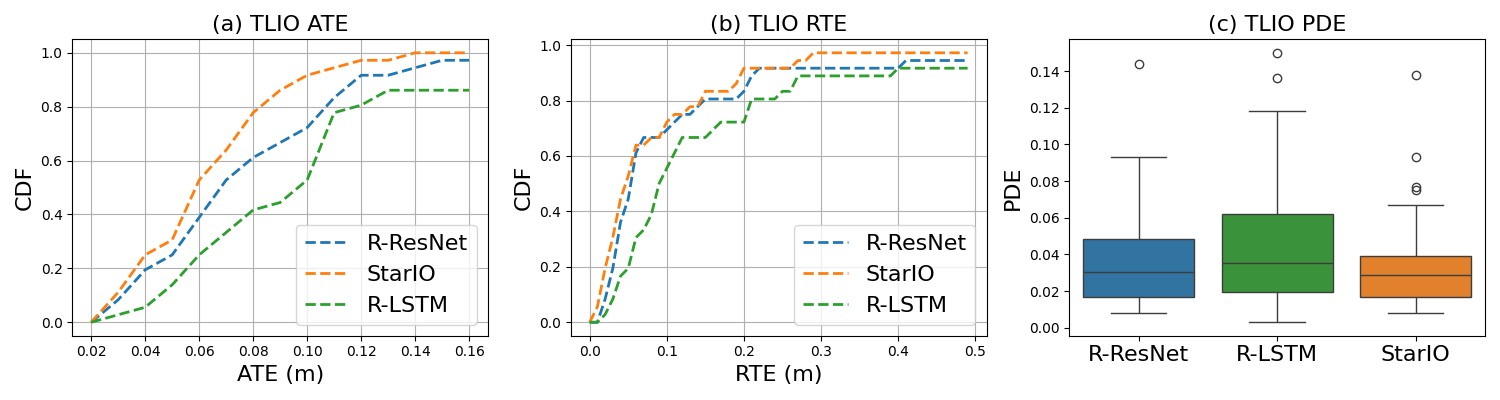}
\caption{Performance comparison of StarIO, R-ResNet, and R-LSTM evaluated on the TLIO dataset. (a) and (b) show the CDF of ATE and RTE, respectively, while subfigure (c) presents a box plot of PDE.}
\label{CDF_and_PDE}
\vspace{-15pt}
\end{figure*}

This operation implicitly maps the measurements into a nonlinear feature space with an approximate dimensionality of $\frac{M(M+1)}{2} \approx \left( \frac{M}{\sqrt{2}} \right)^2$, which holds when $M \gg 2$. Among these components, only $M$ terms are linearly related to the input features $\mathbf{X_{t}}$, while the remaining terms represent higher-order nonlinear interactions—effectively analogous to a kernel function mapping. When stacking $n$ such layers, the resulting feature dimensionality increases to $\left( \frac{M}{\sqrt{2}} \right)^{2n}$, enabling expressive nonlinear modeling even with shallow networks. Therefore, the Star Operation introduces a complementary nonlinearity by modeling feature interactions in a higher-dimensional space, augmenting the effect of conventional activation functions.

This process can be easily extended to handle measurement data over the entire time window $\mathbf{X} \in \mathbb{R}^{C \times L}$, yielding the output $\mathbf{X}_{\text{star}} \in \mathbb{R}^{M \times L}$.

\textit{2) Dual-Wing Attention Mechanism.} Attention mechanisms offer superior capabilities for modeling global relationships. However, standard Transformer self-attention suffers from quadratic computational complexity with respect to measurement sequence length, making it inefficient for long sequences\cite{TKSA}. To address this issue, we propose a Dual-Wing Attention mechanism that captures global motion dynamics along both the temporal and channel dimensions in the high-dimensional implicit nonlinear feature space, while maintaining low computational overhead\cite{MCA}.

Focusing on the channel dimension, given high-dimensional features $\mathbf{X}_{\text{star}}$, we first apply adaptive average pooling (AvgPool) and adaptive standard deviation pooling (StdPool) along the temporal scale to extract contextual statistics along the channel dimension. These are combined using learnable parameters $\mu$ and $\xi$:
\begin{equation}
\mathbf{X'}_{c} = \mu \cdot \mathbf{X}_{\text{star}}^{\text{Avg}} + \xi \cdot \mathbf{X}_{\text{star}}^{\text{Std}} \in \mathbb{R}^{M \times 1}
\end{equation}
where $\mathbf{X}_{\text{star}}^{\text{Avg}}$ and $\mathbf{X}_{\text{star}}^{\text{Std}}$ represent the features obtained via AvgPool and StdPool, respectively. Subsequently, a 1D convolution is applied to $\mathbf{X'}_{c}$ to generate the adaptive channel-wise weights $W_{c} \in \mathbb{R}^{M \times 1} $.

Similarly, by transposing $\mathbf{X}_{\text{star}}$, the same process is repeated along the temporal dimension to obtain $W_{t} \in \mathbb{R}^{L \times 1}$.

Finally, we integrate the attention weights from both dimensions to produce the Dual-Wing representation:
\begin{equation}
\mathbf{X}_{\text{DWS}} = W_{c} \otimes \mathbf{X}_{\text{star}} + (W_{t} \otimes \mathbf{X}_{\text{star}}^T)^T \in \mathbb{R}^{M \times L}
\end{equation}
where $\otimes$ represents channel-wise broadcasting multiplication.
This mechanism effectively captures bidirectional contextual motion information in IMU signals at minimal computational overhead, resulting in substantial improvements in localization accuracy compared to baseline methods.

\textbf{Multi-Scale Gated Convolution Unit (MSGCU).}
In standard Transformer architectures, the attention module is typically followed by a MLP consisting of linear layers and nonlinear activations to enhance the model's representational capacity \cite{VIT}. However, in our architecture the nonlinear representational capacity has already been effectively realized by DWSB. We therefore focus on multi-scale context modeling to aggregate rich contextual information across diverse receptive fields.

We propose the Multi-Scale Gated Channel Unit (MSGCU), which injects contextual motion information into the value branch for inter-channel dependency modeling and incorporates a local dynamic gating mechanism to capture fine-grained feature variations. The MSGCU fuses local and global cues within a unified channel-attention module and comprises two principal branches: the \textbf{value branch} and the \textbf{gating branch}.

\begin{figure*}[!t]
\centering
\captionsetup{aboveskip=2pt,font=small}
\includegraphics[width=1.0\textwidth]{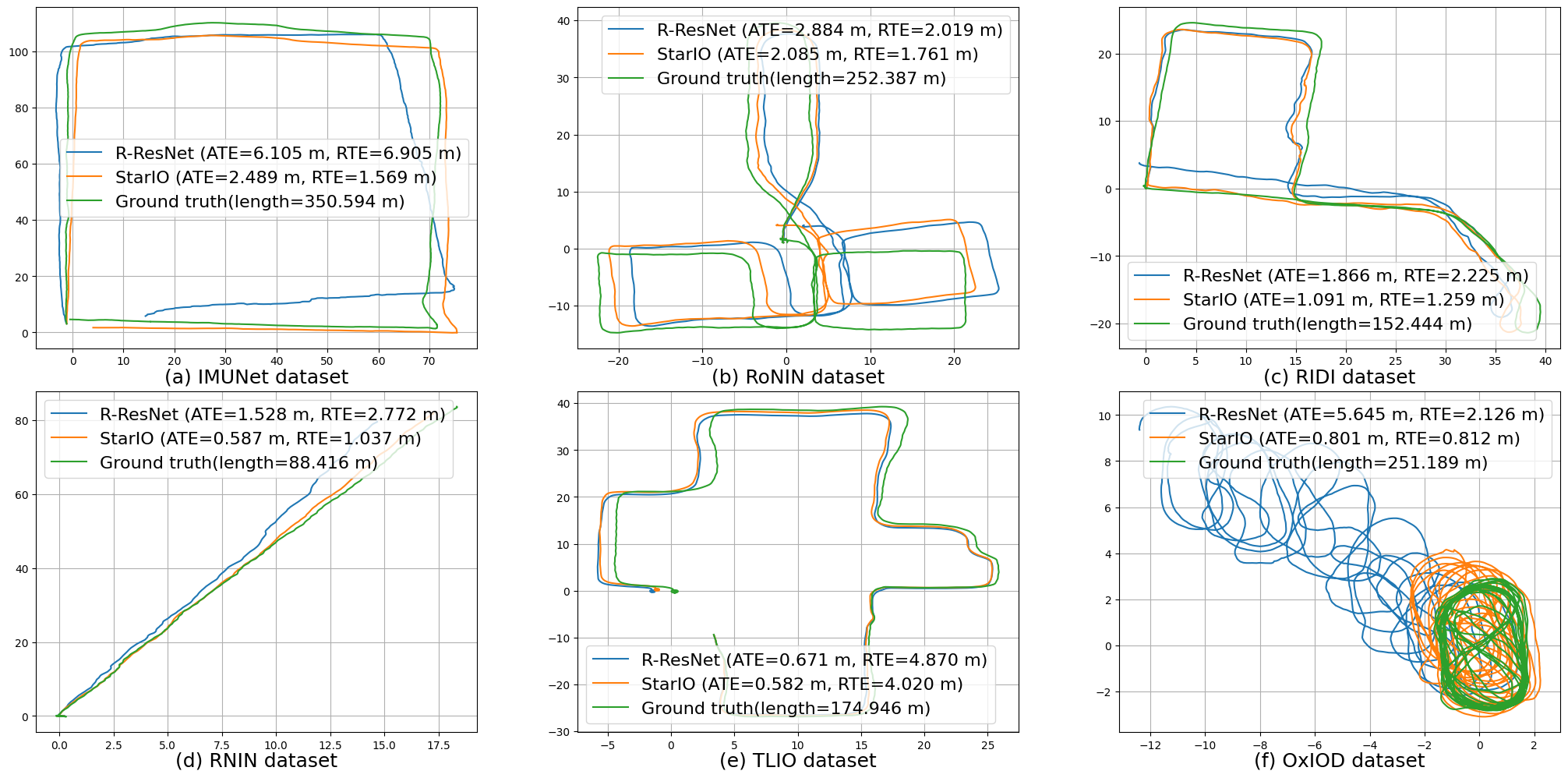}
\caption{Sample trajectories from six test datasets. The results compare the performance of the proposed StarIO model against the R-ResNet baseline. Values in parentheses indicate the ATE, RTE, and trajectory length, all measured in meters.}
\label{combined_plot}
\vspace{-15pt}
\end{figure*}

\textit{1) Value branch.} A depthwise 1-D convolution with kernel size $3$ aggregates local motion-variation features. This branch captures subtle variations across neighboring channels and enables flexible modulation based on local dependencies \cite{TransNeXt}. Such a mechanism is particularly well-suited to IMU signal data, which often exhibit local continuity in motion patterns.

\textit{2) Gating branch.} Inspired by the Squeeze-and-Excitation module\cite{SE}, this branch models global inter-channel dependencies. It first applies global average pooling to the input feature and then employs a two-layer fully connected network to explicitly model channel-wise interactions. A softmax activation is used to produce adaptive selection weights that modulate the local motion-variation features.

These two branches capture channel interactions at local and global scales, respectively. The gating output serves as a gating factor to modulate the aggregated local features from the value branch, thereby achieving multi-scale gated fusion. The resulting channel-attention mechanism not only improves inter-channel discriminability but also enhances adaptability to local perturbations, significantly improving robustness and generalization under complex motion dynamics.

\section{Experiments and Analysis}
\subsection{Experimental Settings}
\textbf{Datasets.} Extensive experiments were conducted on several publicly available inertial localization datasets, including RIDI\cite{RIDI}, OxIOD\cite{Oxiod}, RoNIN\cite{RoNIN}, RNIN\cite{RNIN-VIO}, IMUNet\cite{IMUNet}, and TLIO\cite{TLIO}. All datasets were randomly split into training, validation, and testing sets at an 8:1:1 ratio.

\textbf{Baseline.} Numerous studies indicate that data-driven IO methods achieve significantly higher accuracy than Newtonian-based IO\cite{CTIN,iMOT,NILoc}. Therefore, we focus on comparing against data-driven IO approaches. The baseline models include R-ResNet\cite{ResNet} and R-LSTM\cite{LSTM} introduced in RoNIN\cite{RoNIN}; the neural networks proposed in RNIN\cite{RNIN-VIO} and TLIO\cite{TLIO}; and lightweight networks such as DeepILs\cite{DeepILS}, along with I-IMUNet\cite{IMUNet}, I-MobileNet\cite{Mobilenets}, I-MnasNet\cite{Mnasnet}, and I-EfficientNetB0\cite{Efficientnet} (all proposed in IMUNet\cite{IMUNet}).

\textbf{Metrics.} To comprehensively evaluate localization performance, we use Absolute Trajectory Error (ATE)\cite{IDOL}, Relative Trajectory Error (RTE)\cite{IDOL}, and Position Drift Error (PDE)\cite{CTIN} as the primary metrics. All of these metrics are positively correlated with localization error, where lower values indicate better performance. Trajectory visualizations are also provided to further illustrate the localization accuracy of each method.

\textbf{Implementation Details.} The network configuration is set to $[N_1, N_2, N_3, N_4] = [2, 2, 2, 2]$, yielding a total depth comparable to ResNet-18. All experiments were implemented using PyTorch 2.5.1 and CUDA 12.2. The models were trained using the Adam optimizer, with a maximum of 100 epochs, an initial learning rate of $10^{-4}$, and automatic termination triggered when the learning rate decayed to $10^{-6}$ to prevent overfitting. All training and inference were conducted on an NVIDIA RTX 3090 (24GB) GPU. Other configurations are kept consistent with TLIO\cite{TLIO}.

\subsection{Comparisons with Baselines}
\textbf{Quantitative Comparison.} Table~\ref{results} presents the comparison of ATE and RTE performance on six benchmark datasets. Overall, StarIO outperforms the existing open-source baselines under most testing conditions.
On the RNIN dataset, which includes complex motion patterns such as walking, running, stair climbing, and random shaking under high-noise conditions\cite{RNIN-VIO}, StarIO achieves an ATE of 1.455 m, outperforming RNIN (ATE: 1.605 m). Similarly, on the IMUNet dataset, which involves challenging outdoor scenarios with significant disturbances\cite{IMUNet}, StarIO achieves an ATE of 4.964 m, surpassing I-IMUNet (ATE: 9.130 m). These improvements are attributed to StarIO's ability to map inertial data into a high-dimensional nonlinear feature space and to effectively model complex dynamic motions through multi-scale feature fusion.

\begin{table}[!tbp]
\centering
\captionsetup{aboveskip=2pt,font=small}
\scriptsize  
\renewcommand{\arraystretch}{0.9}  
\caption{The ablation study results on six dataset. The best result is highlighted in bold.}
\label{ablation_study}
\begin{tabular*}{\linewidth}{@{\extracolsep{\fill}} llccc}
\hline
\noalign{\vskip 2pt}  
\multirow{2}{*}{Dataset} & \multirow{2}{*}{Metric} 
& R-ResNet & \begin{tabular}[c]{@{}c@{}}StarIO\\(w/o MSGCU)\end{tabular} & StarIO \\\noalign{\vskip 2pt}  
\cline{3-5}
\noalign{\vskip 2pt}  
& & \multicolumn{3}{c}{Performance (meters)} \\  
\noalign{\vskip 2pt}  
\hline

\noalign{\vskip 2pt}  
\multirow{2}{*}{RoNIN}   & ATE                     & 4.165  & 4.471 & \textbf{3.948} \\
                         & RTE                     & 3.724  & 3.572 & \textbf{3.554} \\ 
\multirow{2}{*}{RIDI}    & ATE                     & 2.080  & 2.118 & \textbf{2.033} \\
                         & RTE                     & 2.383  & 2.446 & \textbf{2.274} \\ 
\multirow{2}{*}{IMUNet}  & ATE                     & 6.428  & 6.577 & \textbf{4.964} \\
                         & RTE                     & 4.802  & 4.601 & \textbf{3.815} \\
\multirow{2}{*}{TLIO}    & ATE                     & 1.333  & 1.254 & \textbf{1.108} \\
                         & RTE                     & 3.705  & 3.525 & \textbf{3.145} \\ 
\multirow{2}{*}{RNIN}    & ATE                     & 1.844  & \textbf{1.250} & 1.455 \\
                         & RTE                     & 1.836  & \textbf{1.213} & 1.370 \\
\multirow{2}{*}{OxIOD}   & ATE                     & 4.336  & 1.502 & \textbf{1.484} \\
                         & RTE                     & 1.595  & 0.878 & \textbf{0.856} \\  
\noalign{\vskip 2pt}  
\hline

\noalign{\vskip 2pt}  
\multicolumn{2}{@{}l}{Parameters (M)} & 4.635 & \textbf{2.252} & 2.762\\
\multicolumn{2}{@{}l}{FLOPs (M)} & 38.252& \textbf{21.555} & 25.122 \\ 
\multicolumn{2}{@{}l}{Peak Memory (MB)} & 22.835& \textbf{9.800} & 20.458 \\  
\noalign{\vskip 2pt}  
\hline
\end{tabular*}
\vspace{-15pt}
\end{table}
    
Notably, StarIO has the second-lowest parameters (2.762M), the third-lowest peak GPU memory usage (20.458MB), and the third-lowest FLOPs (25.122M), comparable in scale to DeepILS. Yet, it achieves significantly lower ATE and RTE, and substantially outperforms existing methods in most scenarios---striking a favorable balance between accuracy and lightweight design.

Additionally, we conducted a visual analysis of the error distribution. Due to space limitations, only the results on the largest dataset (TLIO) are presented. As shown in Fig.~\ref{CDF_and_PDE}, the cumulative distribution function (CDF) of trajectory errors demonstrates that StarIO outperforms R-ResNet and R-LSTM across all accuracy levels. In Fig.~\ref{CDF_and_PDE}(a) and (b), the orange curve (corresponding to StarIO) consistently occupies the upper-left region of the plot, indicating that 80\% of test samples achieve an ATE below 0.08 m (\(P(X<0.08)=0.8\)), whereas R-ResNet and R-LSTM yield values closer to 0.1 m for the same percentage. Furthermore, the boxplot in Fig.~\ref{CDF_and_PDE}(c) shows that StarIO attains the lowest median and smallest dispersion in PDE, further validating its stability and advantage in high-precision localization tasks.

\textbf{Qualitative Visualization.} 
To intuitively illustrate the localization performance, we visualize reconstructed trajectories for several representative test samples in Figure~\ref{combined_plot}, comparing StarIO with R-ResNet. In relatively simple scenarios (e.g., Fig.~\ref{combined_plot}(d)), both models accurately reconstruct the trajectory. However, in more complex cases shown in Fig.~\ref{combined_plot}(a), (b), (c), (e), and (f), R-ResNet exhibits noticeable drift, particularly in sequences involving frequent turns or loops where heading errors accumulate rapidly. In contrast, StarIO more closely matches the ground-truth trajectory, demonstrating superior modeling of complex motion patterns and better capture of long-term dependencies. This further confirms its reliability and practicality for real-world applications.

\subsection{Ablation Studies}
To assess the individual and combined contributions of each module in the proposed architecture, we designed two ablation configurations: full StarIO and StarIO without MSGCU (StarIO (w/o MSGCU)). We conducted systematic experiments across multiple datasets, and Table~\ref{ablation_study} summarizes the results.
StarIO (w/o MSGCU) has 2.252 M parameters---slightly less than half of R-ResNet's 4.635 M---yet achieves localization accuracy comparable to that of R-ResNet. Adding MSGCU increases StarIO's parameter count by only 0.510 M and yields the lowest localization error in most cases.
In summary, the ablation studies validate the individual effectiveness of DWSB and MSGCU as well as their complementarity.

\section{Conclusion}
We present StarIO, a lightweight architecture that enhances the modeling of complex nonlinear motion patterns while maintaining a small parameter footprint. Extensive experiments on multiple publicly available pedestrian-inertial datasets demonstrate the effectiveness of StarIO. However, the applicability of the proposed method to non-pedestrian platforms—such as ground vehicles, unmanned aerial vehicles (UAVs), or mobile robots---remains an open question and warrants further investigation.

\bibliographystyle{IEEEtran}
\bibliography{ref}

\end{document}